\documentclass[sigconf]{acmart}

\AtBeginDocument{%
  }

\copyrightyear{2022}
\acmYear{2022}
\setcopyright{acmcopyright}\acmConference[KDD '22]{Proceedings of the 28th ACM
SIGKDD Conference on Knowledge Discovery and Data Mining}{August 14--18,
2022}{Washington, DC, USA}
\acmBooktitle{Proceedings of the 28th ACM SIGKDD Conference on Knowledge
Discovery and Data Mining (KDD '22), August 14--18, 2022, Washington, DC, USA}
\acmPrice{15.00}
\acmDOI{10.1145/3534678.3539063}
\acmISBN{978-1-4503-9385-0/22/08}

\usepackage{natbib}
\usepackage{bm}
\usepackage{algorithm}
\usepackage{caption}
\usepackage{subcaption}
\usepackage{enumitem}
\usepackage{graphicx}
\usepackage{algorithmic}

\usepackage{tikz}
\usetikzlibrary{tikzmark}
\tikzset{mycircled/.style={circle,draw,inner sep=0.1em,line width=0.04em}}
\usepackage[export]{adjustbox}
\usepackage{multirow}

\begin{document}

\title{A Meta Reinforcement Learning Approach for Predictive Autoscaling in the Cloud}

\author{Siqiao Xue$^*$,Chao Qu$^{*,+}$,Xiaoming Shi, Cong Liao, Shiyi Zhu, Xiaoyu Tan, Lintao Ma, Shiyu Wang, Shijun Wang$^+$, Yun Hu, Lei Lei, Yangfei Zheng, Jianguo Li, James Zhang}
\email{ {siqiao.xsq,peter.sxm,liaocong.lc,zhushiyi.zsy,yulin.txy,lintao.mlt,weiming.wsy,shiyu.wang}@antgroup.com}
\email{{huyun.h,jason.ll,yangfei.zyf,lijg.zero,james.z}@antgroup.com} \email{+{chaoqu.technion,sjwang05}@gmail.com}
\affiliation{%
  \institution{Ant Group}
  \city{Hangzhou}
  \country{China}
}

\renewcommand{\shortauthors}{Xue and Qu, et al.}

\begin{abstract}
  Predictive autoscaling (autoscaling with workload forecasting) is an important mechanism that supports autonomous adjustment of computing resources in accordance with fluctuating workload demands in the Cloud. In recent works, 
Reinforcement Learning (RL) has been introduced as a promising approach to learn the resource management policies to guide the scaling actions under the dynamic and uncertain cloud environment. However, RL methods face the following challenges in steering predictive autoscaling, such as lack of accuracy in decision-making, inefficient sampling and significant variability in workload patterns that may cause policies to fail at test time. To this end, we propose an end-to-end predictive meta model-based RL algorithm, aiming to optimally allocate resource to maintain a stable CPU utilization level, which incorporates a specially-designed deep periodic workload prediction model as the input and embeds the Neural Process \cite{garnelo2018conditional,kim2019attentive} to guide the learning of the optimal scaling actions over numerous application services in the Cloud. Our algorithm not only ensures the predictability and accuracy of the scaling strategy, but also enables the scaling decisions to adapt to the changing workloads with high sample efficiency. Our method has achieved significant performance improvement compared to the existing algorithms and has been  deployed online at Alipay, supporting the autoscaling
of applications for the world-leading payment platform.  
\end{abstract}



\maketitle
\def\thefootnote{*}\footnotetext{These authors contributed equally to this work}
\def\thefootnote{\arabic{footnote}}
\section{Introduction}\label{section:introduction}



One of the key characteristics of operating in the Cloud is autoscaling \footnote{Please see Appendix \ref{section:appendix_background}- \ref{section:appendix_background_2} for a detailed introduction on autoscaling.}, which elastically scales the resources \emph{horizontally} (the number of virtual machines (VMs) assigned is changed) or \emph{vertically} (the CPU and memory reservations are adjusted), to match the changing workload. According to the timing of scaling, the autoscaling strategies can be divided into responsive and predictive strategies. Compared to the responsive ones, predictive strategies forecast the workloads and prepare the resources \emph{in advance} to meet the future demands, therefore yielding better scaling timeliness \cite{asarsa20} and becoming popular in industrial practises \cite{azure_book,aws_book}.



\begin{figure}
     \centering
     \begin{subfigure}[b]{0.5\textwidth}
         \centering
         \includegraphics[width=\textwidth]{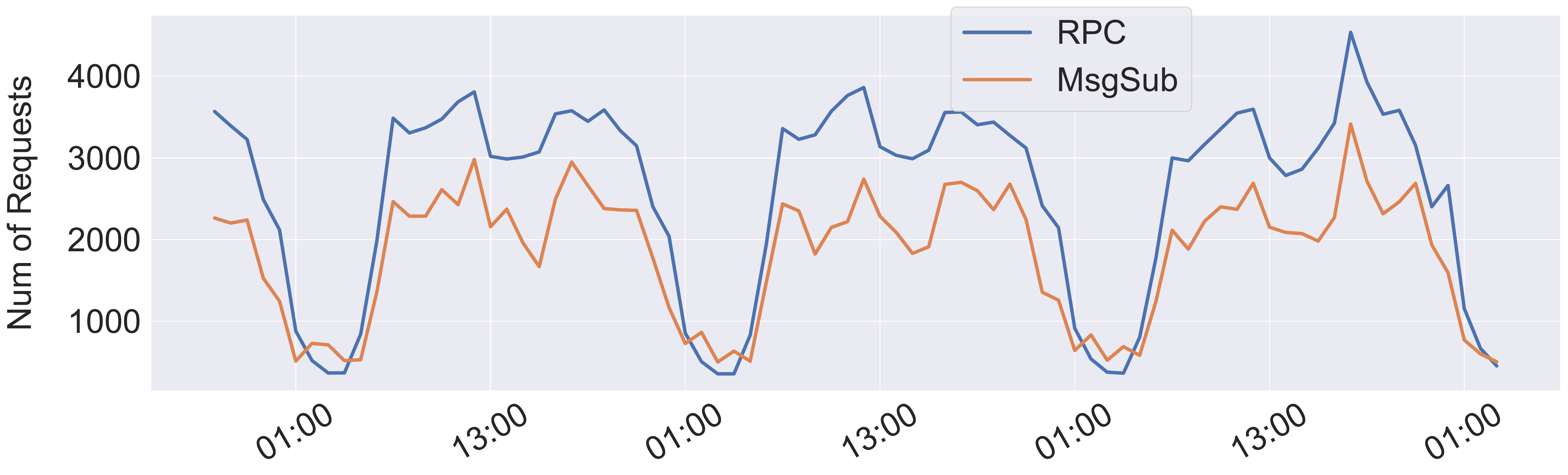}
         \caption{The RPC and MsgSub traffic of a cloud application.}
         \label{fig:workload}
     \end{subfigure}
     \vfill
     \begin{subfigure}[b]{0.5\textwidth}
         \centering
         \includegraphics[width=\textwidth]{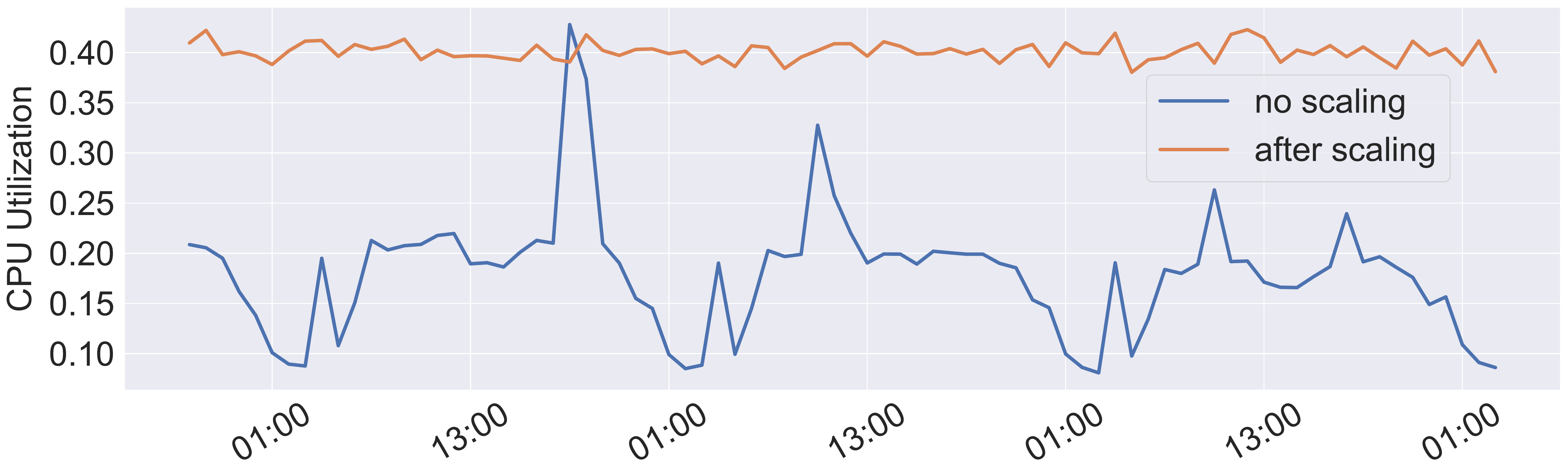}
         \caption{The CPU utilization of the same application.}
         \label{fig:cup_utils}
     \end{subfigure}
        \caption{The multidimensional workload and CPU utilization of a cloud application from July 29th to Aug 1st, 2021.}
        \label{fig:workload_cput_utils}
\end{figure}

In this paper, we focus on building the \textbf{predictive horizontal scaling} strategies at the Cloud of Alipay, the world's leading digital payment platform, to ensure this large-scale system meets its stringent service level objectives (SLOs) \footnote{Please see Appendix \ref{section:appendix_alipay} for a detailed description of the cloud system.}. The system consists of over 3000 running applications/services \footnote{In this paper, we interchangeably use these two terms.} on over 1 million VMs. The operator allocates VMs to applications based on performance indicators (e.g., CPU utilization) dependent on workloads. The workload of applications in Alipay is mainly driven by the traffic of several subtypes, e.g, Remote Procedure Calls (RPC), Message Subscription (MsgSub), etc, and we formulate it as a \textbf{multi-dimensional vector} throughout the paper. Figure \ref{fig:workload} illustrates the evolution of two subtypes of the workload of an online application and Figure \ref{fig:cup_utils} shows that its CPU utilization fluctuates with the workload. Without the scaling, the operator usually allocates resources based on the peak CPU utilization, which produces notably wastage because most of the time, the utilization is far below the peak. Therefore, we aim at timely adjusting the number of VMs according to the needs of workload to \textbf{keep the  CPU utilization of applications running stably at the desired target level} to maximize resource savings.




\setlength{\belowcaptionskip}{-2pt}
\setlength{\belowcaptionskip}{-10pt}

Considering the Cloud is a dynamic and uncertain environment, Reinforcement Learning (RL) serves as a good candidate for autoscaling since it is capable of learning transparent (with no human intervention), dynamic (no static plans) and adaptable (constantly updated) resource management policies to execute scaling actions. Recently, several RL-based methods have been proposed \cite{drlcloud18,rlpas19,asarsa20,fuzzy16,firm2020}, achieving excellent performance in resource saving. However, the existing RL-based methods face several challenges: (i) The prediction of workloads relies on classical time series models (e.g., ARIMA), which have been proven to have limited learning ability compared to the deep learning models \cite{transformer2019,salinas2019deepar, informer2021}; (ii) Most of existing RL algorithms for autoscaling are model-free, which involves prohibitively risky and costly operations in the cloud because they require numerous and direct interactions with the online environment during training;  (iii) The variability in the performance of VMs across different applications is neglected. Previous works either naively think all VMs perform identically or cope with this heterogeneity by modeling them separately, but neglecting the commonalities across them.



\noindent\textbf{Contributions:} In this paper, we propose a novel RL-based predictive autoscaling approach: 
\begin{itemize}[leftmargin=*]
    \item We develop a deep attentive periodic model for multi-dimensional multi-horizon workload prediction, which provides high-precision and reliable workload information for scaling.
    \item We employ a meta-learning model to train a dynamic prior of the map from the workload to CPU utilization, with rapid adaptation to the changing environment, embedded to guide the learning of optimal scaling actions over thousands of online applications. The meta model-based RL algorithm enables safe and data-efficient learning.
    \item To the best of our knowledge, our approach is the first fully differentiable RL-based predictive autoscaling strategy, which has been successfully deployed to support autoscaling at Alipay, the world-leading payment platform.
\end{itemize}



\section{Preliminaries}\label{section:problem_set_up}

\setlength{\belowcaptionskip}{0pt}
\begin{figure}
     \centering
     \begin{subfigure}[b]{0.5\textwidth}
         \centering
         \includegraphics[width=\textwidth]{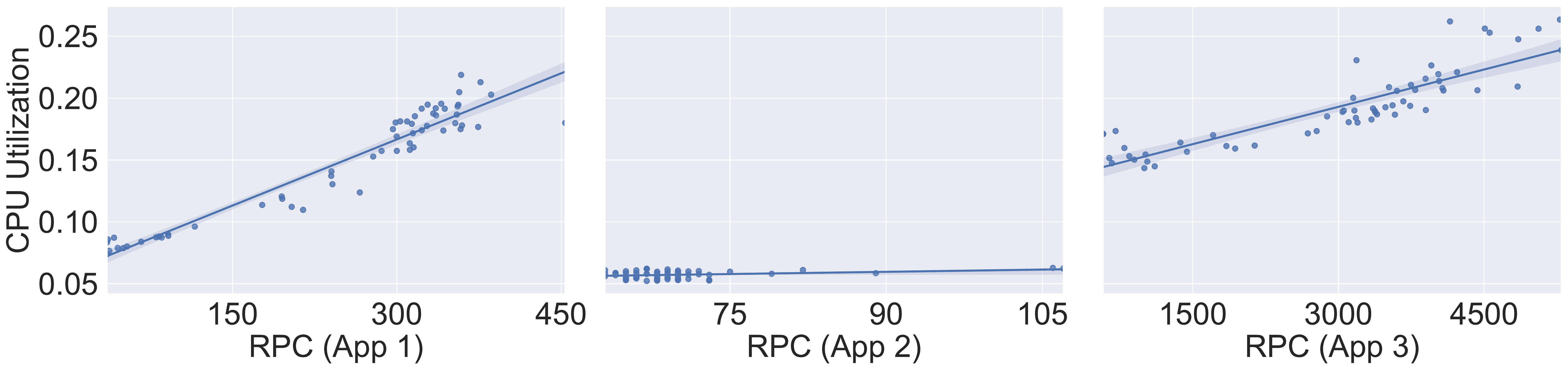}
         \caption{The fitted correlation between RPC traffic and CPU Utilization.}
         \label{fig:cpu_correl}
     \end{subfigure}
     \vfill
     \begin{subfigure}[b]{0.5\textwidth}
         \centering
         \includegraphics[width=\textwidth]{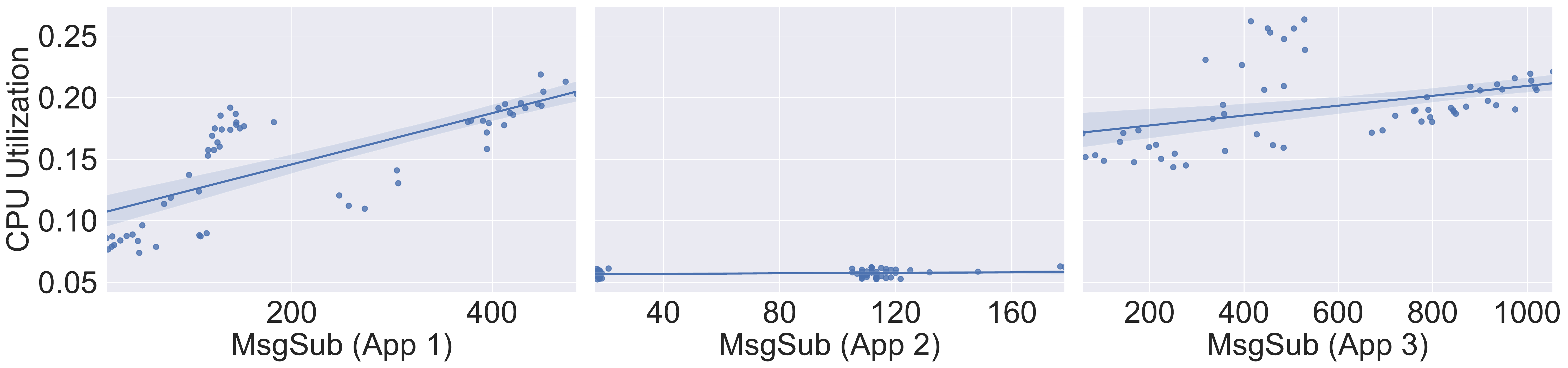}
         \caption{The fitted correlation between MsgSub traffic and CPU Utilization.}
         \label{fig:cpu_correl_2}
     \end{subfigure}
        \caption{The correlation between two subtypes of the workload and CPU utilization of three randomly chosen applications from July 29th to Aug 1st, 2021.}
        \label{fig:cpu_correl_factor}
\end{figure}
\setlength{\belowcaptionskip}{-12pt}

In this section, we firstly present three key insights on the problem as background and motivations and then briefly introduce \textbf{Neural Process} and \textbf{Markov Decision Process} that are building blocks of our proposed approach.

\subsection{Background and Characterization}
\noindent \textbf{Insight 1: The workload patterns usually have a complex composition of various periodicity.} 
The workloads of the most applications at Alipay Cloud are driven by repeatable business behavior (e.g., daily payment during rush hours) with occasional interventions from the platform (e.g., online marketing campaign), and hence usually exhibit a composition of periodicity with abrupt changes, as shown in Figure \ref{fig:workload}. Many existing works use either classical regression techniques \cite{asarsa20} or simple neural networks \cite{Shahin2016,workpred2012} to forecast the workload, which are ineffective in capturing either inherent periodicity or complex temporal dependencies. We resort to deep time series models that have achieved notable success recently \cite{transformer2019,informer2021}.


\noindent \textbf{Insight 2: The workload has heterogeneous impact on CPU utilization.}
As illustrated in Figure \ref{fig:cpu_correl_factor}, the heterogeneity exists in two perspectives: (i) The mapping from workload to CPU utilization varies for different applications; (ii) For the same application, the subtypes of workload have diverse correlation with the CPU utilization. A naive solution is to train a model to learn such mapping for each application, which may have high performance but suffers from unaffordable time and maintenance cost. By defining the learning of the mapping for an application as a \emph{task}, we are motivated to apply the meta-learning techniques \cite{finn2018probabilistic} to train a universal model for all the tasks which exploits commonalities and differences across tasks simultaneously.

\noindent \textbf{Insight 3: Finding optimal resources given CPU utilization estimation forms a dynamic decision process.} The ultimate goal of our approach is to decide accurate resources allocation (VMs) for the application according to the estimation of CPU utilization. The relationship between resources and CPU utilization is complex and adjusting the resources usually incur certain costs in the Cloud (e.g., engineering cost when switching VMs for applications). We resort to RL to find such optimal numbers of VMs while minimizing cost over the long term. Noted that model-based RL is more reliable than model-free methods for large scale Cloud systems because it samples efficiently and effectively avoids the potential risk caused by direct interactions between the scaling model and the online environment during training.



\subsection{Neural Processes and Meta learning}\label{section:neural_processes}
 By taking the \textbf{meta-learning framework}, Neural Process (NP) learns to learn a regression method that maps a context set of observed input-output $(x_i,y_i)$ to a distribution over regression function  \cite{garnelo2018conditional}. Each function models the distribution of the output given an input, conditioned on the context. In particular, condition on  observed \textbf{\underline{C}ontexts} $(x_C,y_C):=(x_i,y_i)_{i\in C}$ each function models \textbf{\underline{T}argets} $(x_T,y_T):=(x_i,y_i)_{i\in T}$, which is $p(y_T|x_T, x_C,y_C)$. The latent version of NP includes a \emph{global} latent variable $z$ to account for uncertainty in the predictions of $y_T$ for a given context: 
 \begin{eqnarray}
     p(y_T|x_T,x_C,y_C):= \int p(y_T|x_T,r_C,z) q(z|x_C,y_C) dz
 \end{eqnarray}
 with $r_C:=r(x_C,y_C)$ where $r$ is a deterministic function that aggregates $(x_C,y_C)$ into a finite dimensional representation with permutation invariance in $C$. The parameters of the encoder $q(z|s_C)$ and the decoder $p(y_T|x_T,r_C,z)$ are learned by maximising the ELBO:
\begin{eqnarray}\label{equ:anp_elbo}
    \log p(y_T|x_T,x_C,y_C) & \geq \mathbb{E}_{q(z|s_T)} [\log p(y_T|x_T,r_C,z) ] \nonumber  \\
    &- D_{kl} (q(z|s_T)||q(z|s_C) ).
\end{eqnarray}
NP can be interpreted as a \emph{meta learning} approach \cite{kim2019attentive,garnelo2018conditional,singh2019sequential}, since the probabilistic representation $z$ captures the current uncertainty over the task, allowing the network to explore in new tasks in a similarly structured manner.  

%

\subsection{Markov Decision Process}
Markov Decision Process (MDP) is described by a 5-tuple ($\mathcal{S}, \mathcal{A}, r, p, \gamma$): $\mathcal{S}$ is the  state space, $\mathcal{A}$ is the action space, $p$ is the transition probability, $r$ is the expected reward, and $\gamma\in [0,1)$ is the discount factor \cite{sutton1998introduction}. That is, for $s\in \mathcal{S}$ and $a\in \mathcal{A}$, $r(s,a)$ is the expected reward, $p(s'|s,a)$ is the probability to reach the state $s'$. A policy is used to select actions in the MDP. In general, the policy is stochastic and denoted by $\pi$, where $\pi(a_t|s_t)$ is the conditional probability density at $a_t$ associated with the policy. The state value evaluated on policy $\pi$ can be represented by $V^\pi(s)= \mathbb{E}_{\pi} [\sum_{t=0}^{\infty} \gamma^t r(s_t,a_t)| s_0=s]$ on immediate reward return $r$ with discount factor $\gamma\in (0,1)$ along the horizon $t$.  The agent aims to seek a policy that maximizes the long term return. In model-based RL, the agent uses a predictive model of the world to ask questions of the form “what will happen if I do action \emph{a}?” to choose the best policy. In the alternative model-free approach, the modeling step is bypassed altogether in favor of learning a control policy directly through the interaction with the environment. In general, the model-based RL is more sample-efficient than the model-free counterpart \cite{chua2018deep}.




\setlength{\belowcaptionskip}{5pt}
\begin{table}[h]
\begin{center}
\begin{tabular}{ l|l  } 
 Symbol  & Description\\
 \hline
 $\mathcal{I}$& The set of applications \\
 $L,H \in \mathbb{R}_{+}$ & The lengths of historical and forecast windows\\
 ID $\in \mathbb{R}_+$  & The unique identifier of each application  \\
 $\bm{x}_{t} \in \mathbb{R}^d$ & $d$-dim workload of an application at $t$ \\ 
 $\bm{u}_{t} \in \mathbb{R}^2$ & $2$-dim time-based covariate at $t$ \\ 
 $c_{t} \in \mathbb{R}_{+}$ & CPU utilization of an application at $t$  \\ 
 $\hat{c}_t \in \mathbb{R}_{+}$ & Predicted CPU utilization of an application at $t$ \\
 $c_{[t,t+1)} \in \mathbb{R}_{+}$ & Average CPU of an application at $[t,t+1)$\\
 \hline
 $l_{t} \in \mathbb{R}_{+}$ &  The number of allocated VMs\\
 $a_{t} \in [-0.5, 2] $  & The adjustment rate of \# allocated VMs
 \\
  $\bm{s}_t \in \mathbb{R}^{d_m}$& The state of MDP for an application at $t$\\
  \hline
  $\bm{\bar{x}}_{t} \in \mathbb{R}^d $ & The unit workload of an application  at $t$\\ 
   $\bm{\bar{x}'}_{t} \in \mathbb{R}^d $ & The unit workload of an application\\
   & after the adjustment at t\\ 
 $\bar{\bm{X}}_t \in \mathbb{R}^{d+2}$  & $\bar{\bm{X}}_t:= (\bm{u}_t, \bar{\bm{x}}_t)$\\  
 $\bar{\bm{X}}'_t \in \mathbb{R}^{d+2}$ &  $\bar{\bm{X}}'_t:= (\bm{u}_t, \bar{\bm{x}'}_t)$
\end{tabular}
\end{center}
\caption {Table of Notations} \label{tab:Notation} 
\end{table}
\setlength{\belowcaptionskip}{-12pt}

\section{Problem Setup}

 Given the historical workload of an application $\bm{x}_{t-L:t}$ at time $t$, we aim to find the optimal VM allocations $a_{t+1:t+H}$ over the future period $H$ to make the CPU Utilization running stably at a target level ( e.g., in Figure \ref{fig:cup_utils} the target CPU utilization is $40\%$). In addition, we want to build a \emph{universal} controller for thousands of heterogeneous tasks which can adapt to the rapidly changing environment or even unseen tasks, guaranteeing the flexibility and the robustness of the controller in the real industrial system. To satisfy the demanding requests, we propose one reasonable assumption to simplify the problem 
that is generally satisfied in practise: 

\noindent\textbf{Assumption 1:} The workload of an application are evenly allocated in VMs, e.g., if the workload is represented as a 2-dim vector of traffic $(RPC, MsgSub)=(100,50)$ running on $5$ VMs, then the unit workload per VM is $(20, 10)$.


We leverage the assumption to build the latent dynamic model in model-based RL. The notation used throughout the paper is in Table \ref{tab:Notation}. In general, \textbf{we use the bar over the alphabet to denote the unit value}, e.g., the unit workload $\bar{x}_t$. The superscript $'$ stands for the value after the adjustment.


\setlength{\belowcaptionskip}{5pt}
\begin{figure*}
     \centering
     \begin{subfigure}[b]{\textwidth}
         \centering
         \includegraphics[width=\textwidth]{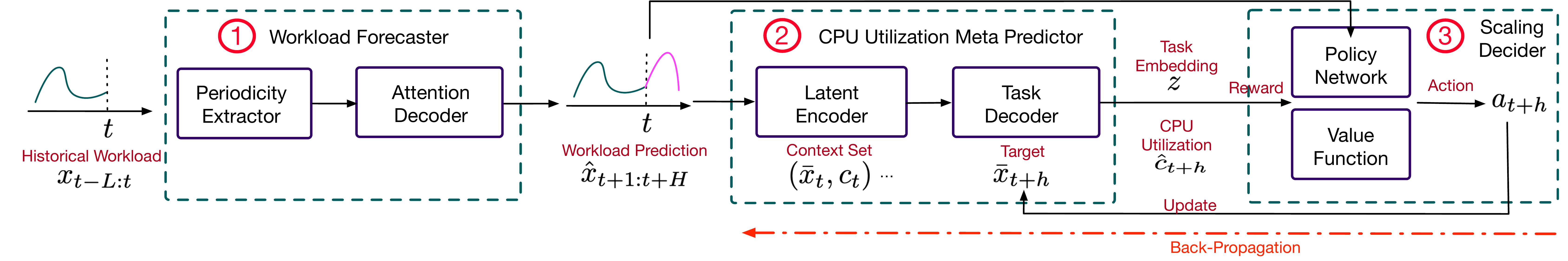}
         \caption{Synergy diagram of Workload Forecaster, CPU Utilization Meta Predictor, and Scaling Decider.}
         \label{fig:model_pipe}
     \end{subfigure}
     \vfill
     \begin{subfigure}[b]{0.35\textwidth}
        \centering
         \includegraphics[width=\textwidth]{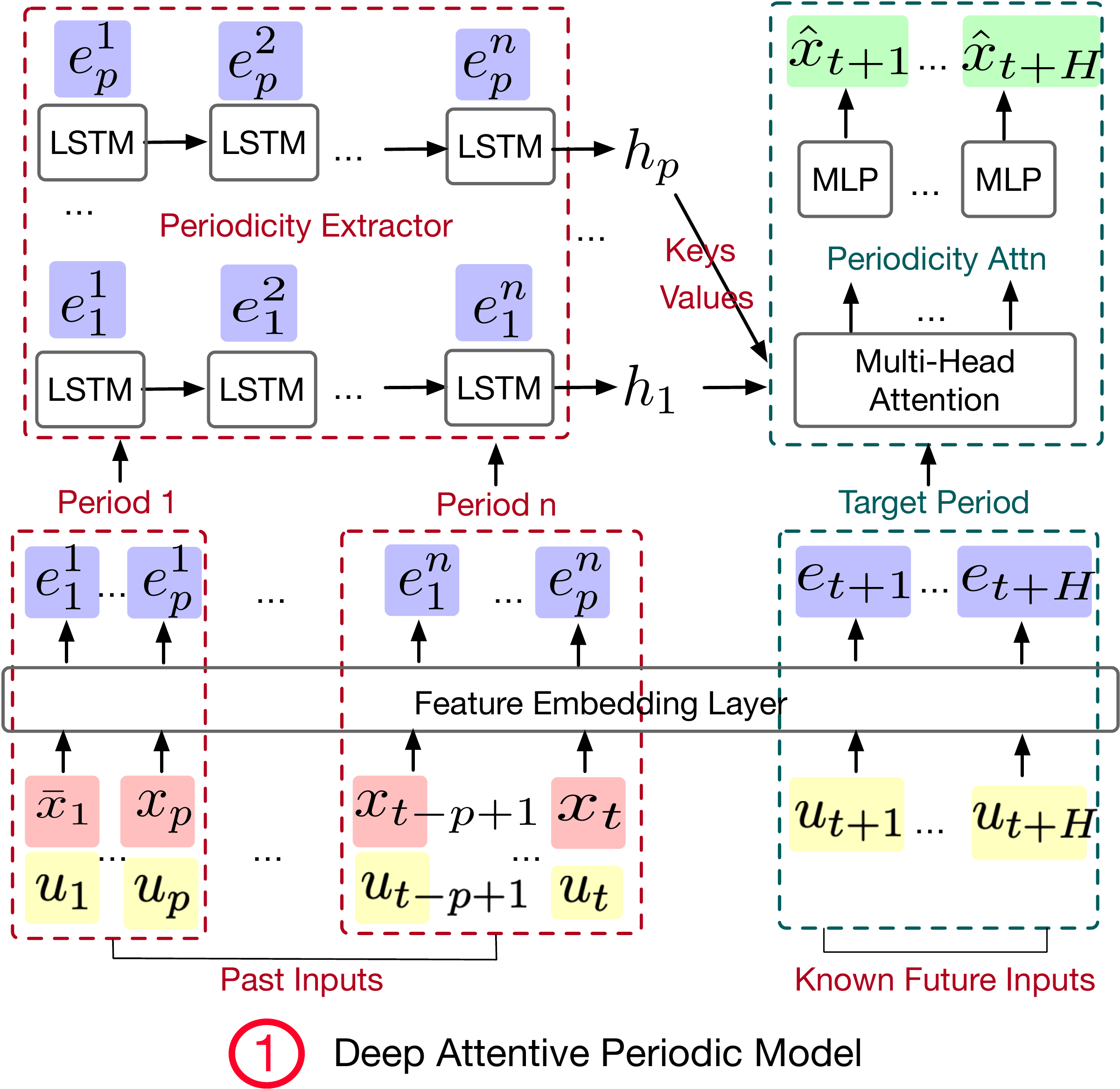}
         \caption{The architecture of Workload Forecaster.}
         \label{fig:model_damp}
     \end{subfigure}
     \hfill
     \quad
     \begin{subfigure}[b]{0.35\textwidth}
        \centering
         \includegraphics[width=\textwidth]{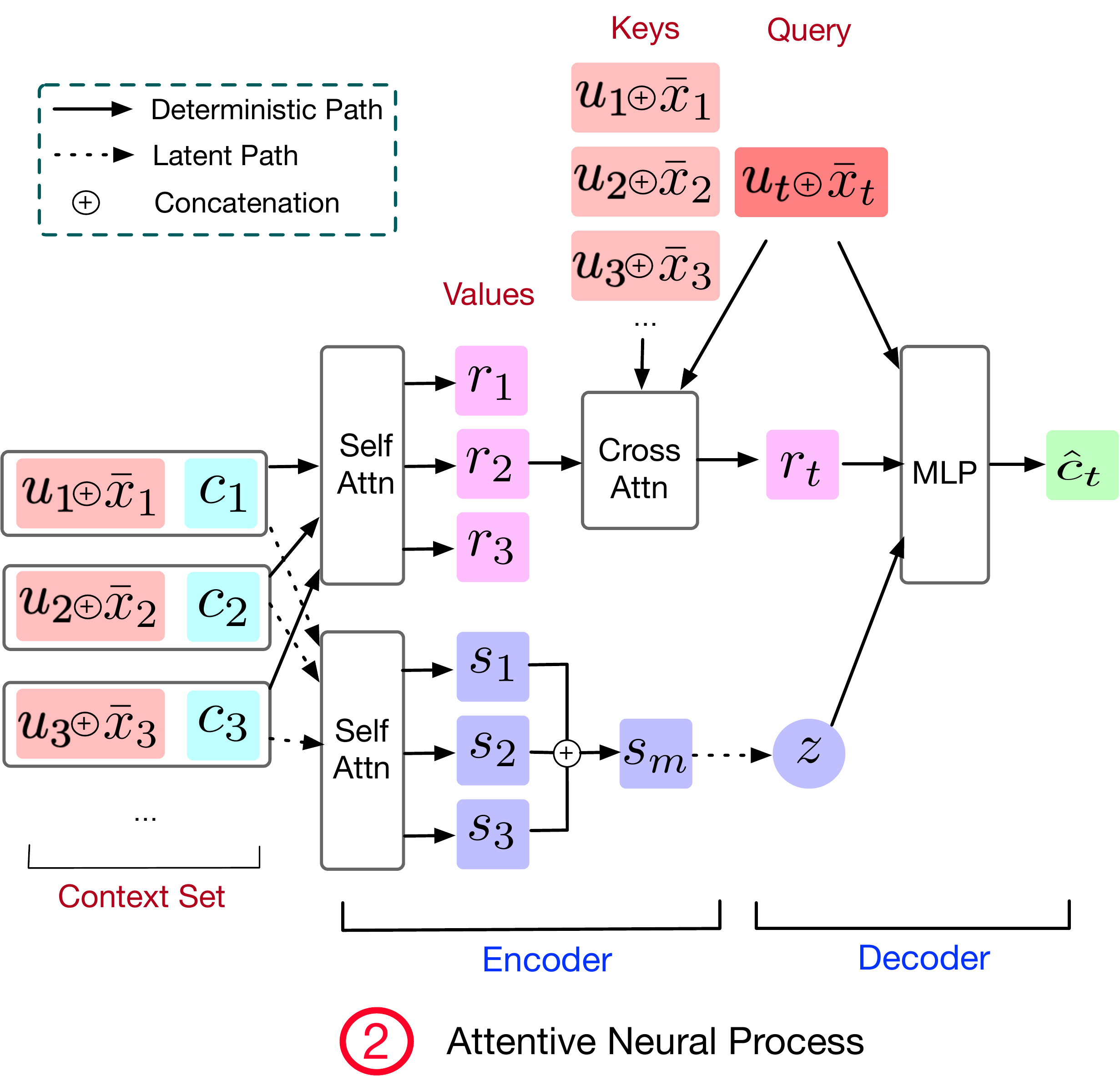}
         \caption{The architecture of CPU Utilization Meta Predictor}
         \label{fig:model_np}
     \end{subfigure}
     \hfill
     \quad
     \begin{subfigure}[b]{0.25\textwidth}
        \centering
         \includegraphics[width=\textwidth]{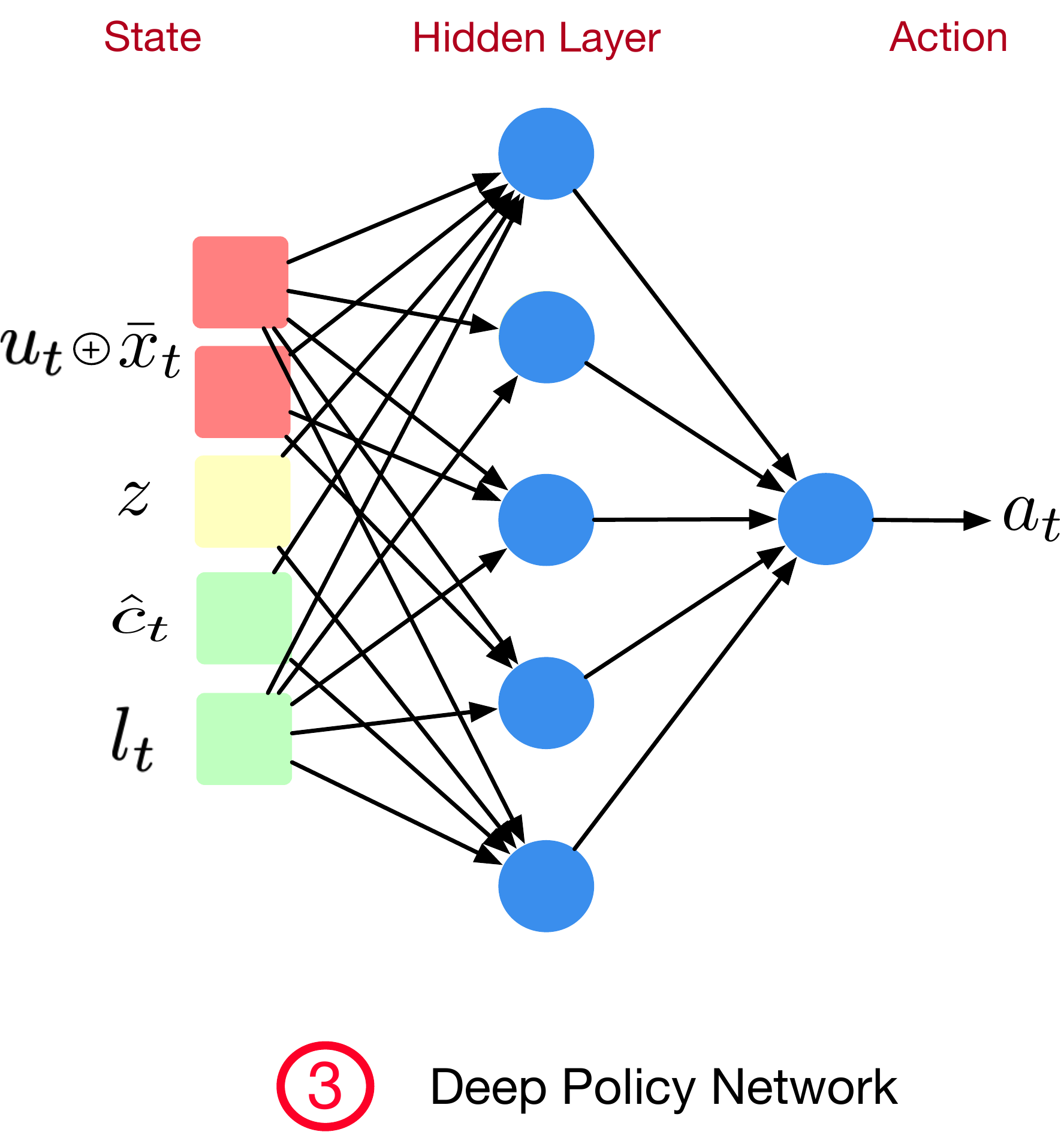}
         \caption{The architecture of Scaling Decider.}
         \label{fig:model_rl}
     \end{subfigure}
        \caption{The end-to-end predictive autoscaling framework.}
        \label{fig:model_arch}
\end{figure*}
\setlength{\belowcaptionskip}{-12pt}

\section{End-to-End Learning to Autoscale}
Three primary components comprise our approach:
\begin{enumerate}[leftmargin=*]
    \item A deep attentive periodic time series model that that holds sufficient capacity to capture complex periodicity of workload patterns discussed in \emph{Insight 1}.
    \item A neural process model that meta-learns the map from the workload to the CPU utilization of applications, which addresses the heterogeneous effect discussed in \emph{Insight 2}. 
    \item An MDP with the above two models embedded to seek the optimal numbers of VMs so that the CPU utilization is kept at a target level, resulting in a model-based RL that avoids the potential deficiencies described in \emph{Insight 3}.
\end{enumerate}
The three components are marked as $\tikzmarknode[mycircled,draw=red,text=red]{t1}{1},\tikzmarknode[mycircled,draw=red,text=red]{t1}{2},\tikzmarknode[mycircled,draw=red,text=red]{t1}{3}$ in Figure \ref{fig:model_arch} and described below respectively. Different from the previous works \cite{asarsa20, predmicro2018}, the components are all amenable to auto-differentiation, which effectively avoid issues of convergence stability and data-efficiency, constituting a single global autoscaling strategy.


\subsection{Stage 1: Predict Workload Pattern via Deep Attentive Periodic Time Series Model}
\label{section:ts_model}
In our approach, workload prediction is needed to estimate
the incoming workload of the applications for future periods. As mentioned in \emph{Insight 1}, most of the previous works utilize classical time series models and the application of advanced deep models are less well studied. Nonetheless, instead of directly applying the state-of-the arts, such as \textit{Informer} \cite{informer2021} and \textit{ConvTransformer} \cite{transformer2019}, we propose Deep Attentive Periodic Model (\textbf{DAPM}) with two distinctive characteristics for workload predictions: 
\begin{itemize}[leftmargin=*]
    \item A lightweight \emph{periodicity 
extractor} that captures inherent seasonality of the workload, where the data exhibits constant patterns of rises and falls, as seen in Figure \ref{fig:workload}.
    \item A \emph{periodicity attention} module that learns complicated periodic dependencies of the workload. 
\end{itemize}

We reduce the problem to learning a universal prediction model for all applications
\begin{eqnarray}
    \bm{x}_{t+1:t+H} = f_{\theta}\left( \bm{x}_{t-L:t}, \bm{u}_{t-L:t+H} \right),
    \label{eqn:ts_pred}
\end{eqnarray}
where $\theta$ is the parameters of the function $f$ and $\bm{u}_{t-L:t+H}$ is a set of $2$-dim covariates $\bm{u}_t=\{u_{t, d}, u_{t, h}\}$ assumed to be known over the entire time period: day-of-the-week $u_{t,d}$ and hour-of-the-day $u_{t,h}$. We firstly initialize a feature embedding layer to generate input embedding $\bm{e}_t \in \mathbb{R}^m$ at each timestamp
\begin{eqnarray}
    \bm{e}_t = W^e [\bm{x}_t; f_{\theta_{u}}(\bm{u}_{t})]+b^e,
\end{eqnarray}
where $f$ is an embedding map of the covariates. Secondly, we cluster all the $\bm{e}_i$  into $n$ groups with $p$ vectors in each. Then we construct $n$ LSTM layers with the $i$-th LSTM layer taking $\{\bm{e}^j_i, j=1:n\}$ as the input. The rationale behind is, by setting $p$ to be the seasonality length of the workload, we can utilize LSTM to learn the dynamics of inherent periodical behaviors, e.g., rise and fall pattern every 24 hours, explicitly at each snapshot. 

Finally, we pass the last hidden state of each LSTM layers into the multihead attention \cite{vaswani2017attention} as $K=V=\{\bm{h}_1, ..., \bm{h}_p\}$ with embeddings form known future covariates are taken as Queries $Q=\{\bm{e}_{t+1},...,\bm{e}_{t+H}\}$, which learns a weighted combination of hidden representations of periodicity:
\begin{eqnarray}
\small
    \bm{S}=Multihead(Q,K,V)= softmax(\frac{QW^Q (KW^K)^T}{\sqrt{d}} VW^V),
    \label{eqn:multihead}
\end{eqnarray}
where $W_h^Q \in \mathbb{R}^{d_Q\times d},W_h^K \in \mathbb{R}^{d_K\times d}$ and $W_h^V \in \mathbb{R}^{d_V\times d}$ are learned linear transformation matrices for query, key and value respectively. The \emph{multihead} lies in using different sets of weight $\{W^Q_h, W^K_h,W^V_h\}_{h=1}^H$ to compute a set of attention output $\{\bm{S}_1,...,\bm{S}_H\}$ and the final output of the attention is $\bm{S}=[\bm{S}_1,...,\bm{S}_H] W^O$, which is followed by MLP to produce the multi-step predictions $\bm{\hat{x}}_{t+1:t+H}$. The model is trained by minimizing the RMSE loss between $\bm{x}_{t+1:t+H}$ and $\bm{\hat{x}}_{t+1:t+H}$.



\subsection{Stage 2: Learn the Mapping From Workload to CPU Utilization via Neural Process}\label{section:cpu_utilization}
As discussed in \emph{Insight 2}, the fact that VMs perform heterogeneously \emph{across the applications} motivates \textbf{the use of meta-learning on the uncertainty over tasks of learning the heterogeneous mapping from the unit workload $\bar{\bm{x}}_t:=\frac{\bm{x}_t}{l_t}$ to CPU utilization $c_t$}. We choose Attentive Neural Process (ANP), a state-of-the-art NP from DeepMind \cite{kim2019attentive}. As depicted in Figure \ref{fig:model_np}, ANP 
uses an attention module to encode complex dependencies between the context along with a probabilistic representation vector capturing the global distribution of uncertainty over tasks, helping automate downstream scaling decision-making problem.

Considering the existence of heterogeneity among applications, we introduce the  covariate $\bm{u}_t$ as an auxiliary feature to the model input along with the unit workload: $\bm{X}_t=( \bm{u}_t, \bar{\bm{x}}_t)$. Then we define the context set as $\mathcal{C}=\{(\bm{X}_s, c_s)\}^{ID \in \mathcal{I}}_{s\in [t-L:t]}$ and target set as $\mathcal{T}=\{(\bm{X}_s, c_s)\}^{ID \in \mathcal{I}}_{s\in [t+1:t+H]}$. An input embedding layer is designed to transform $X_t$ into a dense vector before feeding into the model. In particular, we map the context information $(\bar{x}_i, c_i)$ into a finite dimensional representation $r_i$ through a self-attention module. Given the target input, i.e., the query in Figure \ref{fig:model_np}, we do the cross multihead attention, whose formulation is described in Equation \ref{eqn:multihead}, over the key-value pairs to predict the target output $r_t$.  



As depicted in Figure \ref{fig:model_np}, besides this deterministic representation, ANP has a stochastic path, where the output $z$ is a global latent variable to account for uncertainty in the prediction of $c_t$ for a given observed $(\bm{X}_s,c_s)$.  The structure is similar to the deterministic one, but now we aggregate $s_i$ into a single vector $s$ by taking the mean. $z$ is modelled as a factorized Gaussian distribution using the reparameterization trick with its mean and variance \cite{kingma2013auto}.

The encoder $q$ outputs the hidden representation $z$ and the reference $r$ while the decoder $p$ aggregates the information of $r_c,z,X_T$ and outputs the target output $y_t$. ANP is trained by optimizing Equation \ref{equ:anp_elbo} over the context and target sets, with two folds of output:
\begin{itemize}[leftmargin=*]
    \item Predicted CPU utilization $\hat{c}_t$ on the target set \footnote{The input of target set uses ground truths of $\bm{\bar{x}}_s, s\in[t-L:t]$ in training stage while using predicted $\bm{\widehat{\bar{x}}}^i_s, s\in [t+1:t+H]$ in the test stage.}.
    \item A global probabilistic latent representation $z$, which can be seen as the \emph{task embedding} encoding salient information of the task given the context information. Intuitively, it tells the agents what kind of tasks they face. 
\end{itemize}
In the following, we demonstrate how to leverage the learned ANP in the dynamic model of the model-based RL.



\subsection{Stage 3: Autoscale via Meta Model-based RL}
\label{section:rl}
Based on \emph{Insight 3}, we establish the scaling process as a model-based RL algorithm. Given the workload prediction from \emph{Stage 1}, the agent learns to continually scale the number of VMs by interacting with the CPU utilization estimator trained from \emph{Stage 2}. The goal is to keep the CPU utilization stable in the future period. 
\subsubsection{The MDP Formulation}\label{section:MDP}
We define the MDP of the scaling strategy as:
\begin{itemize}[leftmargin=*]
    \item State space $\mathcal{S}$: State $\bm{s}_t:=(\bar{\bm{X}}_t, z, \hat{c}_t, l_t$) is a tuple of unit workload, task embedding $z$, the corresponding estimated CPU utilization, and the number of allocated VMs.
    \item Action space $\mathcal{A}$:  We design the adjustment rates $a_t$, so that the number of VMs allocated $l_t$ typically takes the form of  $l_{t+1} = l_{t}\times (1+a_{t})$ where $a_t$ is the adjustment rate. In practice, we set the lower bound and the upper bound of the $a_t$ to be $-0.5$ and $2$, respectively.
    \item Reward $r_t$: The immediate reward at time step $t$ is defined as  $ r_t:=-( c_{[t,t+1)}-c_{target})^2- \eta (l_{t+1} - l_{t})^2$, which is the weighted sum of two terms: (i) a distance between current CPU utilization and the target one; (ii) a switching cost to penalize the adjustments of VMs, which makes our strategy applicable in the real-world setting. The hyper parameter $\eta$ is a positive constant to balance between the two terms.
    \item Discount factor $\gamma$:  In our case we set to $\gamma = 0.95$.
\end{itemize}
\emph{Assumption 1} indicates $\bar{\bm{x}}'_{t}= \frac{{\bm{x}}_{t}}{l_t(1+a_t)}$ and we use ANP trained in \emph{Stage 2} to predict the CPU utilization $\hat{c}_t$ after the adjustment. 
In particular,  $$\hat{c}'_t = f_{ANP}(\bar{X}_t') = f_{ANP}((\bm{u}_t, \bar{\bm{x}}'_{t} )) = f_{ANP}((\bm{u}_t, \frac{x_t}{l_{t}(1+a_{t})} )).$$ Recall that the first term of reward function is  $-(c_{[t,t+1)} - c_{target})^2$ and in practice,  we use the predicted value $\hat{c}'_t $ to replace $c_{[t,t+1)}$. Then we plug  $l_{t+1} = l_{t}(1+a_{t})$ into the second term, and we have the reward function:
\begin{equation}\label{equ:reward_function}
r_t = -(\hat{c}'_t- c_{target})^2 - \eta (a_{t} l_{t})^2.
\end{equation}
Notice that $r_t$ is a function of $s_t$ and $a_t$, i.e., $r_t = r(s_t,a_t)$. After imposing the action $a_t$, the system arrives at the next state $s_{t+1} = (\bar{\bm{X}}_{t+1},z,\hat{c}_{t+1},l_{t+1} ).$ Regarding the term $\bar{\bm{X}}_{t+1}$ and $ \hat{c}_{t+1}$, we use the predicted workload at $t+1$ obtained from the time series model to estimate $\bar{\bm{X}}_{t+1}$, which is then fed into $f_{ANP}$ to obtain $\hat{c}_{t+1}$.  $z$ is a time invariant term to characterize the property of tasks. $l_{t+1} = l_t(1+a_{t})$ is  the dynamics over the allocated VMs.     To ease the exposition, we denote overall dynamics over the state by  
$$\bm{s}_{t+1} = g(\bm{s}_{t},a_t).$$
\textbf{Key observations}: Both reward and dynamic models are  differentiable w.r.t. the input action $a_t$. In particular, $\hat{c}'_t$ is \emph{differentiable} w.r.t. the action $a_{t}$ through the mapping of $f_{ANP}$ and therefore the $r_t$ is also a differentiable function w.r.t. $a_{t}$.  We will leverage these key properties in the derivation of the following model-based RL algorithm.

\subsubsection{Policy Learning}

The policy $\pi(a_t|s_t)$ is a mapping from the state to the distribution of action. For simplicity, we assume the policy is deterministic and is parameterized by a neural network, i.e., $a_t = \pi_\psi (\bm{s}_t)$, where $\psi$ are the weights of the neural network. Different from the model-free RL algorithm such as Q-learning \cite{van2016deep} and Actor-Critic \cite{sutton1998introduction}, we do not learn the optimal policy from the interactions with environment by TD-learning.  Remind that we have already obtained the dynamic model $s_{t+1} = g(s_t,a_t)$ and reward model $r(s_t,a_t)$. To find the optimal $\psi$ to maximize the long term reward, we embed above two models into the Bellman equation. Recall that Bellman equation for the value function:
\begin{eqnarray}
    V^{\pi}(\bm{s}_t) = \mathbb{E} [r(\bm{s}_t,a_t) + \gamma V^{\pi}(\bm{s}_{t+1})],
    \label{eqn:bellman}
\end{eqnarray}
where the expectation is over the randomness of the dynamic model. We unroll the Bellman equation $H$ steps, which is 
\begin{eqnarray}
    V^{\pi}(\bm{s}_t) = \mathbb{E} [r(\bm{s}_t,a_t) &&+ \gamma r(s_{t+1},a_{t+1})+,...,+\gamma^H r(s_{t+H},a_{t+H}) \nonumber \\
     &&+\gamma^{H+1} V^{\pi}(\bm{s}_{t+H+1})],
\end{eqnarray}
In practice, we discard the  term $\gamma^{H+1} V^{\pi}(s_{t+H+1})$. On one hand, we have the a discount factor $\gamma^{H+1}\ll 1$ and this term has a negligible effect. On the other hand, the length of forecast  window  is $H$ in Section \ref{section:ts_model} and we do not know the dynamics and reward beyond $H$ steps. Therefore, we have a simplified form:
\begin{equation}\label{equ:bellman_rollout}
      V^{\pi}(\bm{s}_t) \approx \mathbb{E} [r(\bm{s}_t,a_t) + \gamma r(s_{t+1},a_{t+1})+,...,+ \gamma^H r(s_{t+H},a_{t+H})]
\end{equation}
We simulate the trajectory from $ t$ to $t+H$ using the dynamic model and current policy $\pi$ as that in \cite{hafner2019dream} to obtain a Monte-Carlo approximation.  In particular, we have $s_{t+1} = g(s_t,\pi_\psi(s_t))$, $s_{t+2} = g(g(s_t,a_t),\pi_{\psi}(s_{t+1})) $ and so on.  The next step is to conduct the policy improvement over the current policy. Remark that the right hand side of  Equation \ref{equ:bellman_rollout} is a differentiable function w.r.t. $\psi$ and we can carry out the gradient ascent to improve the policy by the automatic differentiation in Tensorflow \cite{abadi2016tensorflow}, known as stochastic value gradient \cite{heess2015learning}.
\begin{equation}\label{equ:value_gradient}
    \psi \leftarrow \psi + \frac{\partial V^{\pi}}{\partial \psi}.
\end{equation}
We combine all pieces together to obtain our Meta Model-based Predictive Autoscaling (MMPA), with the pseudocode presented in Algorithm \ref{alg:MMPA}.

\begin{algorithm}
\caption{Learning Algorithm of MMPA}\label{alg:MMPA}
\begin{algorithmic}
\STATE{\bfseries Input:} 1)  The dataset $ \{\bm{x}_{t-L:t}, \bm{x}_{t+1:t+H}\} $ to train DAPM (Section \ref{section:ts_model} ). 2) Context set $\mathcal{C}$ and target set $\mathcal{T}$  to train the ANP Model (Section \ref{section:cpu_utilization} ).

\STATE{\bfseries Pretrain DAPM:} 

Optimize the RMSE loss between $\bm{x_{t+1:t+H}}$ and $\bm{\hat{x}_{t+1:t+H}}$. \\
\STATE{\bfseries Pretrain ANP:} 

For each iteration, we randomly sample tasks from context set $C$ and target set $T$ to train the ANP model by optimizing the ELBO in Equation \ref{equ:anp_elbo}. Finish the training until ANP converges. \\

\STATE{\bfseries Train Policy:}
\FOR{each RL step }
\STATE{\bfseries Unroll the dynamic model and calculate reward:}\\

Using  dynamic model $g$ to simulate the state $s_{t+1}$ to $s_{t+H}$ and corresponding reward $r_{t+1}$ to $r_{t+H} $. Obtain a Monte-Carlo approximation of Equation \ref{equ:bellman_rollout}.

\STATE{\bfseries Update policy:}\\
 Update policy $\psi$ using stochastic value gradient, i.e., Equation \ref{equ:value_gradient}.\\
\ENDFOR
\end{algorithmic}
\end{algorithm}


\section{Experiments}
\subsection{Setup}
\subsubsection{Dataset.} We collected one month 10min-frequency data of the workload and CPU utilization of 50 core applications from the cloud system of Alipay for \textbf{offline evaluation} of the workload and CPU prediction. The workload is constructed as a 7-dimensional vector, containing traffics of Remote Procedure Calls (RPC), Message Subscription (MsgSub), Message Push (MsgPush), External Application (EA), Database Access (DA), Write Buffer (WB) and Page View (PV). The details of dataset could be found in Appendix \ref{sec:dataset}.

The first three weeks are the train period while the last week is the test period. Illustrated in Table \ref{tab:selected_app}, we perform the \textbf{online evaluation} of 5 representative applications for the end-to-end scaling over the period of a week, where we scale the application every $4$ hour based on predicted workload, i.e., the historical and predictive windows are set to $288$ and $24$, respectively. 
\setlength{\belowcaptionskip}{5pt}
\begin{table}[H]
\begin{center}
\begin{tabular}{ |c|c|c|c| } 
\hline
Symbol & Service domain & Dominant traffic \\
\hline 
A1 &  file service &  RPC, EA  \\
A2 &  database &  DA, WB  \\
A3 &  web &  RPC, PV  \\
A4 &  tool &  RPC, DA  \\
A4 &  messaging &  RPC, MsgSub, MsgPush  \\
\hline
\end{tabular}
\end{center}
\caption{Descriptions of 5 sample applications}
\label{tab:selected_app}
\end{table}
\setlength{\belowcaptionskip}{-10pt}


\subsubsection{Baselines and Metrics.} Firstly we compare the workload prediction performance of DAMP with two state-of-the-arts: \textit{Informer} \cite{informer2021} and \textit{ConvTransformer} \cite{transformer2019}. Then we validate NP by comparing its capability of CPU utilization estimation with two  widely-applied classic methods including \emph{LR} and \emph{XGBoost} in the context of autoscaling. Finally, we compare our end-to-end scaling approach against two industrial benchmarks: 
\begin{itemize}[leftmargin=*]
    \item Autopilot \cite{drlcloud18}: a workload-based autoscaling method proposed by Google, which builds the optimal resource configuration by seeking the best matched historical time window to the current window. We implement Autopilot based on its public paper \cite{drlcloud18}.
    \item FIRM \cite{firm2020}: a RL-based autoscaling method, which solves the problem through learning feedback adjustment with the online cloud environment. Specifically, FIRM finds applications with abnormal response time (RT) through SVM-based anomaly detection algorithms and adjusts multiple resources for the service through RL algorithms. We implement FIRM using the author's gitlab code  \emph{https://gitlab.engr.illinois.edu/DEPEND/firm}.
\end{itemize}

We evaluate the workload and CPU utilization prediction by MAE and RMSE. For end-to-end autoscaling, we set the target CPU utilization at $40\%$ and evaluate the performance of the strategy by RCS (Relative CPU Stability rate) with a $2\%$ error, i.e., the percentage of time the CPU utilization is within $40\% \pm 2\%$.

\subsubsection{Implementation of Our Approach.} We implemented our approach based on Python 3.6.3 and Tensorflow 1.13.1. The details of the implementation and sample code \footnote{\emph{https://github.com/iLevyFan/meta$\_$rl$\_$scaling}} can be found in Appendix \ref{sec:implementation}.

\setlength{\belowcaptionskip}{5pt}
\begin{table}[h]
    \begin{subtable}[h]{0.5\textwidth}
        \centering
        \begin{tabular}{ |l|l|l|l| } 
        \hline
        Target & Method & MAE & RMSE \\
        \hline
        \multirow{6}{4em}{Workload} & \multirow{2}{7em}{Informer} & 1.75 & 202.10\\ 
        &  & (0.15) & (19.8)\\ 
        & \multirow{2}{7em}{ConvTransformer} & 1.50 & 172.84 \\ 
        &  & (0.19) & (18.9) \\ 
        & \multirow{2}{7em}{\textbf{Ours}} & \textbf{1.10} & \textbf{112.59} \\ 
        &  & (0.09) & (13.1) \\ 
        \hline
        \multirow{6}{4em}{CPU Utilization} & \multirow{2}{4em}{LR} & 1.40 & 2.48\\ 
        & &(0.15) & (0.24)\\
        & \multirow{2}{4em}{XGBoost} & 1.23 & 1.93 \\ 
        & &(0.14) & (0.20)\\
        & \multirow{2}{4em}{\textbf{Ours}} & \textbf{0.86} & \textbf{1.11} \\ 
        & &(0.09) & (0.11)\\
        \hline
        \end{tabular}
        \caption{Workload and CPU utilization prediction evaluation.}
        \label{tab:exp_cpu_pred_res}
    \end{subtable}
    \vfill
    \begin{subtable}[h]{0.5\textwidth}
        \centering
        \begin{tabular}{ |l|l|l|l|l|l| } 
        \hline
            Method & A1 & A2 & A3 & A4 & A5 \\
            \hline
            \multirow{2}{4em}{Autopilot} & 0.77 & 0.75 & 0.65 & 0.66 & 0.84\\
                       & (0.056) & (0.044) & (0.062) & (0.059) & (0.041)\\
                       \hline
            \multirow{2}{4em}{FIRM} & 0.81 & 0.79 & 0.85 & 0.80 & 0.86 \\
            & (0.086) & (0.084) & (0.069) & (0.077) & (0.061)\\
            \hline
            \multirow{2}{4em}{Ours} & \textbf{0.95} & \textbf{0.93} & \textbf{0.92} & \textbf{0.95} & \textbf{0.91} \\ 
            & (0.076) & (0.094) & (0.083) & (0.099) & (0.071)\\
            \hline
        \end{tabular}
        \caption{Relative CPU stability rate of autoscaling strategy evaluation.}
        \label{tab:exp_scaling_res}
    \end{subtable}
    \begin{subtable}[h]{0.5\textwidth}
        \centering
        \begin{tabular}{ |l|l|l|l|l|l| } 
        \hline
            \# Test Applications & 50 & 100 & 200 & 500 \\
            \hline
            \multirow{2}{4em}{RMSE} & 1.11 & 1.19& 1.31 & 1.33 \\
                 & (0.14) & (0.13) & (0.13) & (0.15) \\
            \hline
        \end{tabular}
        \caption{CPU utilization prediction error over large-size test set.}
        \label{tab:exp_cpu_error_scalability}
    \end{subtable}
\caption{Mean and standard deviation (in bracket) of experiment results on test period}
\end{table}
\setlength{\belowcaptionskip}{-15pt}

\setlength{\belowcaptionskip}{0pt}
\begin{figure}
     \begin{subfigure}[b]{0.5\textwidth}
         \centering
         \includegraphics[width=\textwidth]{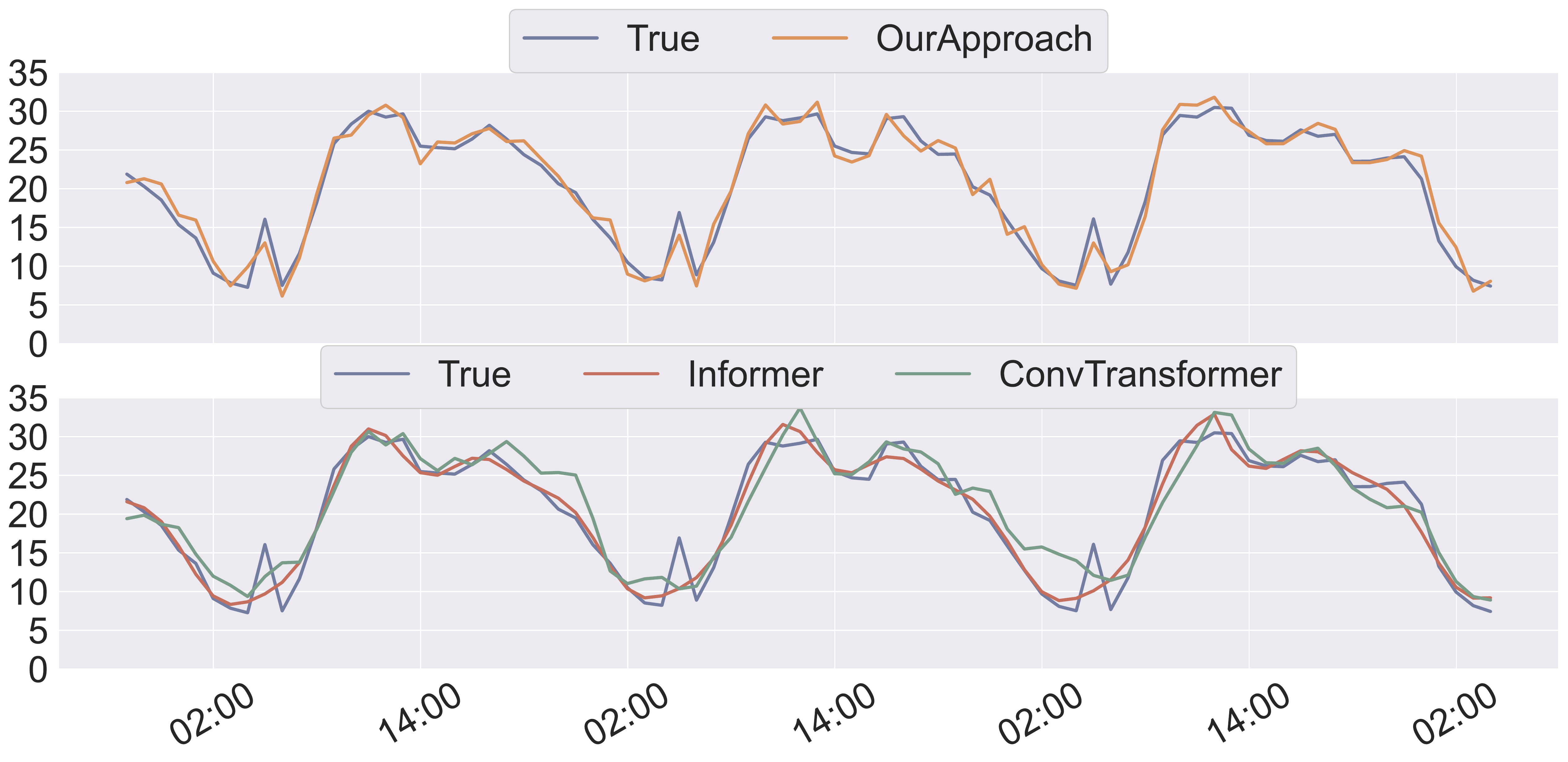}
         \caption{The workload (RPC traffic) forecast of application A1.}
         \label{fig:workload_pred}
     \end{subfigure}
     \vfill
     \begin{subfigure}[b]{0.5\textwidth}
         \includegraphics[width=\textwidth]{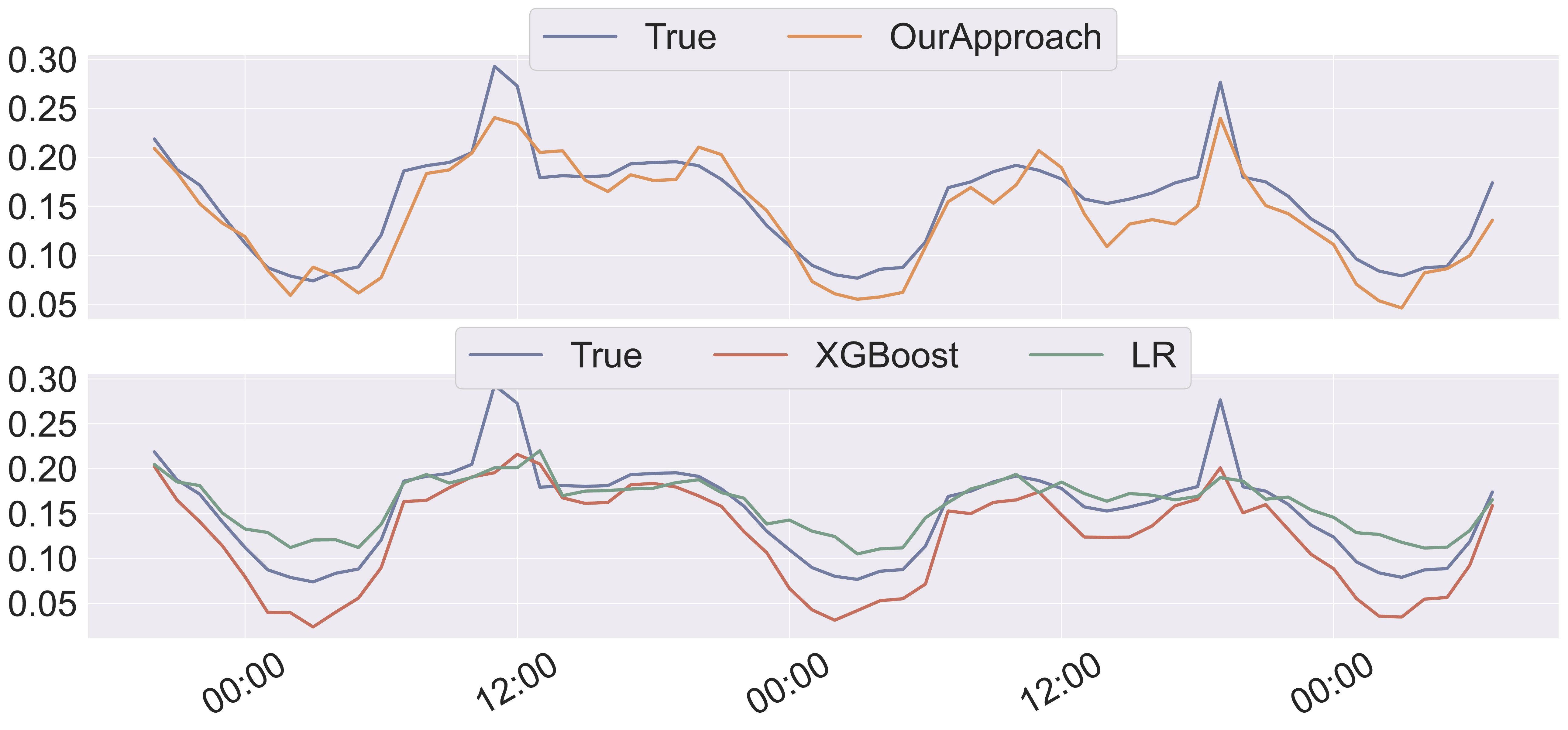}
         \caption{The CPU utilization prediction of application A1.}
         \label{fig:cpu_pred}
     \end{subfigure}
     \vfill
     \begin{subfigure}[b]{0.5\textwidth}
         \includegraphics[width=\textwidth]{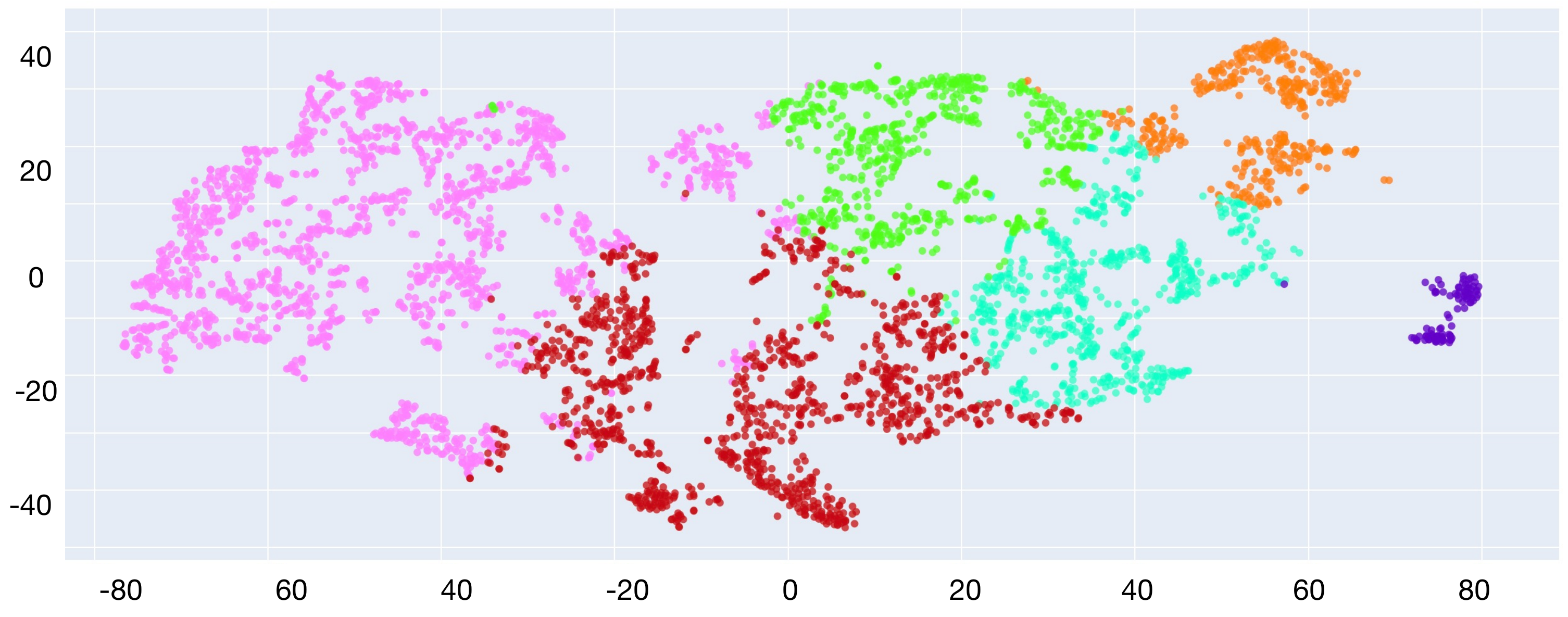}
         \caption{Visualization of dense representations of input $X_t$ in the training set of ANP with t-SNE.}
         \label{fig:np_pred}
     \end{subfigure}
        \caption{Experiment results on workload and CPU utilization prediction.}
        \label{fig:exp_res1}
\end{figure}
\setlength{\belowcaptionskip}{-10pt}

\setlength{\belowcaptionskip}{0pt}
\begin{figure}
     \begin{subfigure}[b]{0.5\textwidth}
         \centering
         \includegraphics[width=\textwidth]{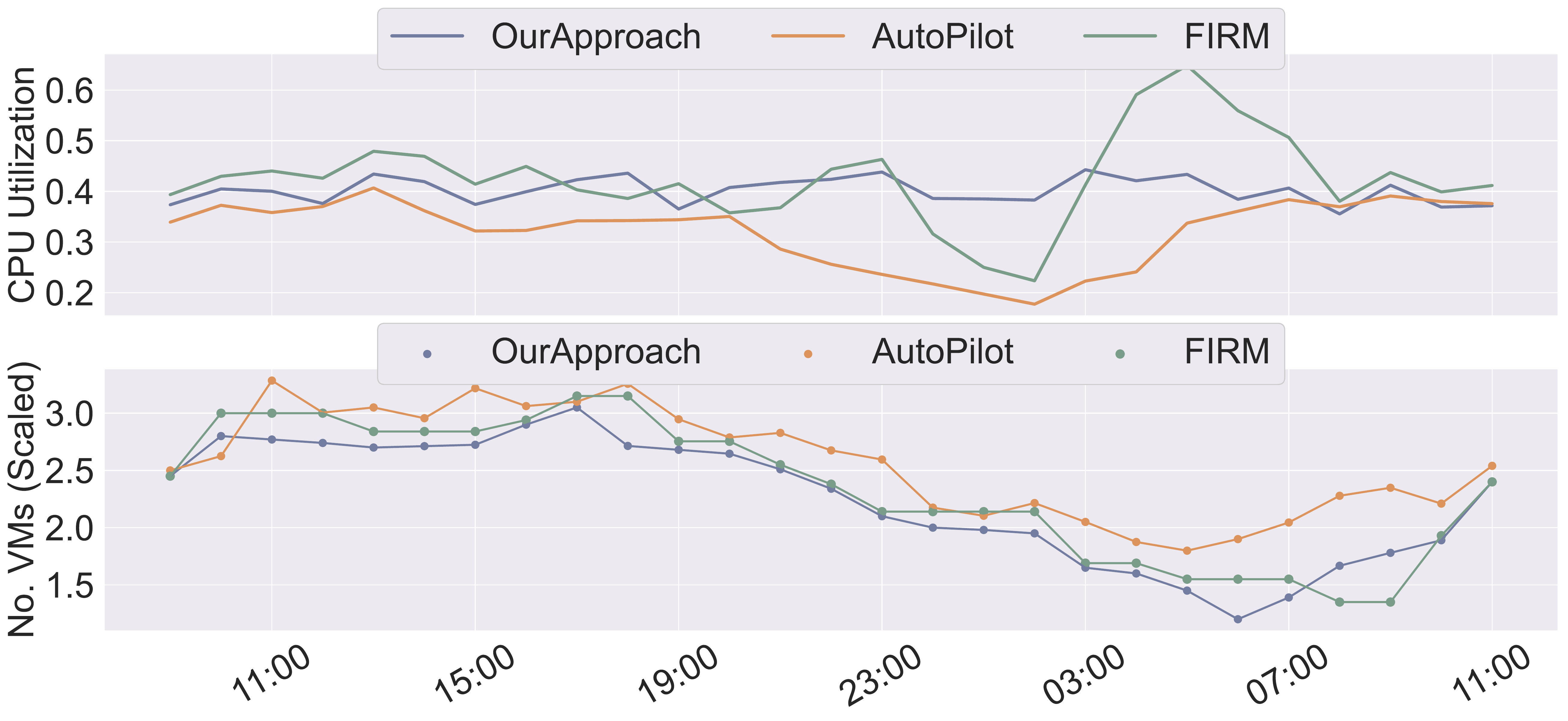}
         \caption{CPU utilization of application A1.}
         \label{fig:cpu_a1}
     \end{subfigure}
     \vfill
     \begin{subfigure}[b]{0.5\textwidth}
         \includegraphics[width=\textwidth]{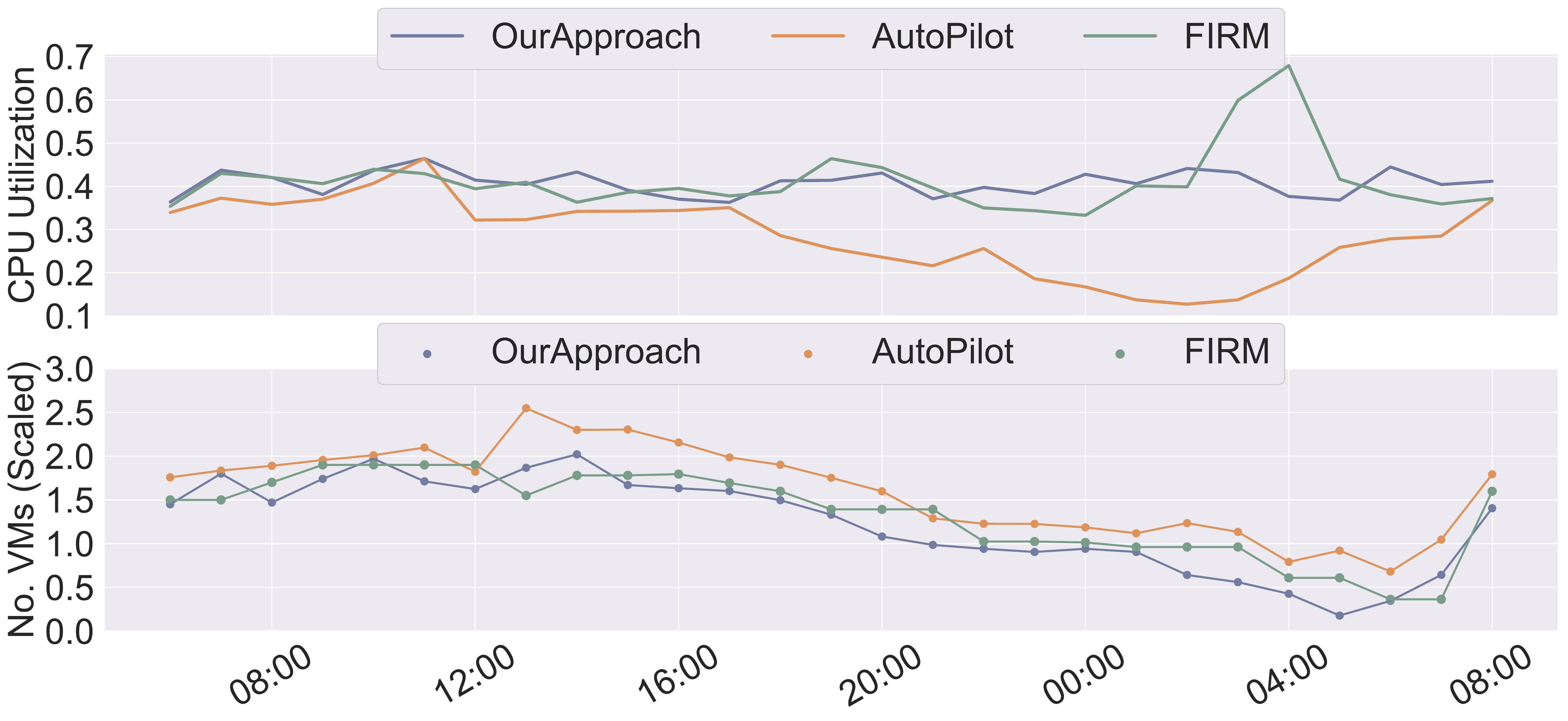}
         \caption{CPU utilization of application A2.}
         \label{fig:cup_a2}
     \end{subfigure}
        \caption{CPU utilization trend of application A1 and A2 on a test period.}
        \label{fig:exp_res2}
\end{figure}
\setlength{\belowcaptionskip}{-15pt}

\subsection{Experiment Results}
\subsubsection{Effectiveness of Workload Prediction.} Generally, Informer and ConvTransformer have failed to capture certain complex temporal patterns of the time series, e.g, in Figure \ref{fig:workload_pred}, change points at 3 am of Application A1 are not well captured. Instead, our approach generates predictions that follows closely in the rise and fall of various periodicity because we employ modules especially designed to better capture composition of temporal patterns and periodicities. As a result, our model achieves a better MSE/RMSE ratio summarized in Table \ref{tab:exp_cpu_pred_res}. 

\subsubsection{Effectiveness of CPU Utilization Prediction} As in Figure \ref{fig:cpu_pred} and Table \ref{tab:exp_cpu_pred_res}, both LR and XGboost perform relatively poor due to their inability to handle complex nonlinear data. 

To validate the effectiveness of the meta-learning in our model, we firstly use t-SNE to plot the dense representations of all training inputs specified in Section \ref{section:cpu_utilization}, which shows $6$ cluster centroids in the latent space, illustrated in Figure \ref{fig:np_pred}, suggesting that when a new CPU utilization prediction task comes, its inputs may be possibly categorized into one of the groups in the latent space so that we can perform the prediction in a similarly structured manner. Then we train our model ANP, the-state-of-art meta model in the domain of CV, and run the prediction on the target set. The results in Table \ref{tab:exp_cpu_pred_res} demonstrate that our approach obtains smaller errors in the test period with around $25\%$ improvement in MSE/RMSE, which more effectively learns the heterogeneous relation between the workload and the CPU utilization.

\subsubsection{Effectiveness of Online Predictive Scaling} Due to space limitation, we only show the scaling results of A1 and A2 in Figure \ref{fig:exp_res2}. The summary of all five services are presented in Table \ref{tab:exp_scaling_res}. As in Figure \ref{fig:exp_res2}, our approach achieves a steady CPU utilization around the target level during the entire day while Autopilot and FIRM have substantial fluctuations, e.g., Application 1 experiences utilization swings (rise and fall ) under FIRM from 23:00-6:00 and is always under target level from 19:00-6:00. 

Overall, our approach obtains \textbf{a higher CPU utilization steady rate and a less total number of VMs}: estimated from Table \ref{tab:exp_scaling_res} and related statistics, the average steady rate of our approach is $19\%$/$10\%$ higher than Autopilot/FIRM and number of VMs of our method is $21\%$/$9\%$ lower than Autopilot/FIRM. We summarize the reasons in three-folds: 
\begin{itemize}[leftmargin=*]
    \item Autopilot and FIRM adjust the resource based on exception detection instead of high-performance predictive techniques, therefore when encountering CPU utilization fluctuations, it takes time to bring it back to the target level by adjusting the numbers of VMs, during which the changing workload patterns may lead to swings in CPU utilization again. 
    \item Our approach, based on effective predictions of workload and CPU utilization, is able to adjust the VMs in advance with minor error, which stabilizes the CPU utilization.
    \item With the embedded meta model of utilization estimation, our agent makes more accurate scaling decisions than that of FIRM trained in a model-free manner. 
\end{itemize}

\subsubsection{Scalability} Our approach has two particular characteristics that help with scalability: 
\begin{itemize}[leftmargin=*]
    \item  We have trained a meta CPU utilization predictor ANP, facilitating fast adaptation to solve new prediction tasks without retraining the whole data again. As seen from Table \ref{tab:exp_cpu_error_scalability}, the RMSE only increases by $20\%$ while the number of predicted applications increase by $10$ times, which proves that our approach can be effectively applied to performing the prediction on a large scale of applications. 
    \item As discussed in Section \ref{section:rl}, our approach forms a fully differentiable scaling strategy that adequately avoid issues of convergence stability and data-efficiency, and is therefore applicable in large scale cloud systems. 
\end{itemize}


\subsubsection{Deployment} Evidenced by the effectiveness and scalibility
, our framework has been deployed in the real-world daily practices of Alipay Cloud. Compared to the \emph{rule-based method} in production, the CPU stability rate has been improved by over $20\%$ with around $50\%$ cloud resources saved.

\section{Related Work}
\label{sec:relatedwork}
The recent works of autoscaling can be categorized into \emph{rule-based} approaches and \emph{learning-based} approaches. In typical rule-based approaches \cite{aws_book,azure_book}, the key goal is to find the threshold that triggers the scaling mechanism, such as a fixed CPU value or an average response time. However such approaches usually require significant human efforts and experience and may fail to respond to changing workload swiftly. The learning-based approaches \cite{Shahin2016,Acampora2017,asarsa20,abdu-2020,adbu2019}, which apply machine learning models to find abnormal states (e.g., CPU utilization is too high) of the system and  optimize the resource, help to address these challenges. For example, \citep{abdu-2020} applies regression trees to model the relationship between the number of machines and response time and then generates the recommended number of machines to avoid service response time over time.


Considering the scaling decision is taken under dynamic and uncertain environment in the online cloud, RL-based learning methods \cite{Arabnejad2016AnAC,rlpas19,drlcloud18,firm2020} have been proposed to model the autoscaling as decision-making problems. \textbf{A distinctive difference} between our method and theirs is that we build a high-performance fully differentiable framework with a meta model-based RL algorithm to perform an end-to-end scaling strategy. It is noted that traditional model-based RL works  \cite{hafner2019dream,heess2015learning,chua2018deep} can not be directly applied because their algorithms only aim to solve a single task, which is impractical in dealing with numerous different applications in the
industrial context. Instead, our RL algorithm is specially tailored for an industrial autoscaling system, incorporating specially-designed modules such as workload forecaster and CPU utilization meta predictor.

\section{Conclusion}
We proposed a novel meta RL-based model, which is, to the best of our knowledge, the first RL-based fully differentiable framework for predictive scaling in the Cloud. Our approach is effective in stabilizing CPU utilization of applications and has been deployed in the real world to support the scaling in the Cloud of a world-leading mobile payment platform, with around $50\%$ resource saved compared to the rule-based method in production.

\bibliographystyle{ACM-Reference-Format}
\bibliography{autoscaling}

\clearpage
\newpage 

\appendix
\section{Appendix}

\subsection{Background on Autoscaling}
\subsubsection{Why Autoscaling}
\label{section:appendix_background}
Autoscaling is a cloud computing feature that enables operators to scale cloud services such as server capacities or VMs up or down automatically, based on defined situations such as traffic or utilization levels. The overall benefit of autoscaling is that it eliminates the need to respond manually in real-time to traffic spikes that merit new resources and instances by automatically changing the active number of VMs.

\begin{figure}[h]
\includegraphics[width=0.4\textwidth]{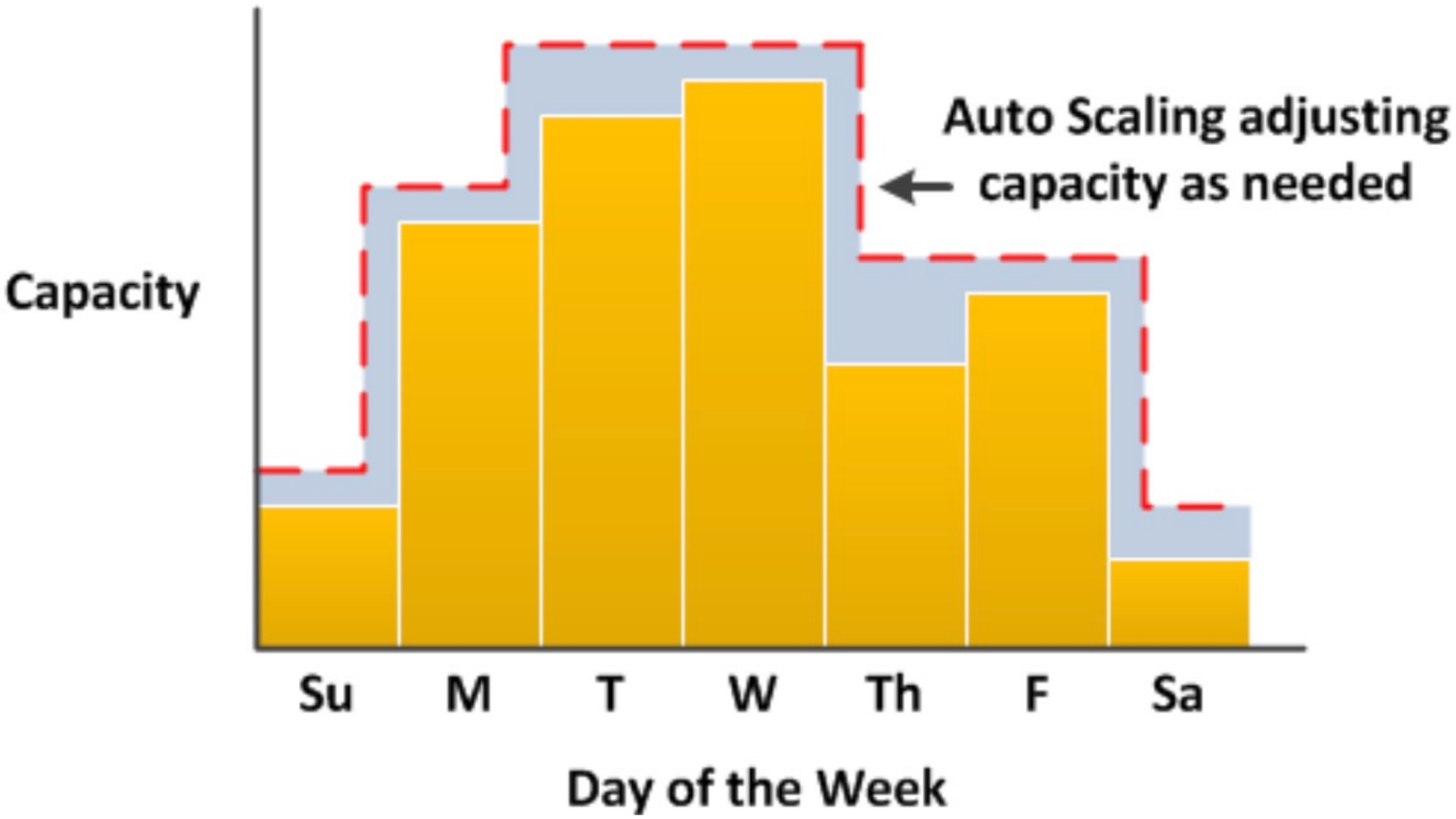}
\caption{An illustration of autoscaling, adopted from Internet.}
\end{figure}

\subsubsection{Why Predictive Autoscaling}
\label{section:appendix_background_2}
Predictive scaling uses certain models to analyze each resource's historical workload and regularly forecasts the future load. Using the forecast, predictive scaling generates scheduled scaling actions to make sure that the resource capacity is available \textbf{before the application needs it}, therefore yielding better scaling timeliness \cite{asarsa20}. In industrial practises \cite{aws_book,azure_book}, predictive scaling works to maintain the utilization at the target value specified by the scaling strategy.


\subsubsection{The Cloud system}
\label{section:appendix_alipay}
Our work is performed in the context of a large-scale production cloud service system from a world-leading online mobile payment provider. The system consists of over 3000 services/applications running on over 1 million VMs while the workload typically has millions of access request per minute. The payment service requires $7 \times 24$ hours availability and the SLO in terms of the success rate for accesses every second is required to be higher than $99.9995\%$. In this paper, we primarily focus on autoscaling, to ensure that this large-scale system meets its stringent SLOs.

\subsection{Experiment Details}
\label{section:appendix_exp}
In this section, we present the details of the experiment deferred from the main text.

\subsubsection{Dataset}\label{sec:dataset}

The workload is defined as a 7-dimensional vector, each of the element corresponds to a subtype of the traffic, which is described below:
\begin{itemize}[leftmargin=*]
    \item Remote Procedure Calls (RPC):  the traffic of external access through the system.
    \item Message Subscription (MsgSub): the traffic of message subscription throughout the system.
    \item Message Push (MsgPush): the traffic of message push throughout the system.
    \item External Application (EA): the traffic of access from external applications.
    \item Database Access (DA): the traffic of database access. 
    \item Write Buffer (WB): the traffic of write buffers throughout the system.
    \item Page View (PV): the traffic caused by user page view.
\end{itemize}

The data does not contain any Personal Identifiable Information (PII), is desensitized, encrypted, is only used for academic research, it does not represent any real business situation. Adequate data protection was carried out during the experiment to prevent the risk of data copy leakage, and the data set was destroyed after the experiment.


\subsubsection{Implementation details}\label{sec:implementation} We implemented our approach based on Python 3.6.3 and Tensorflow 1.13.1. The size of all hidden states in DAPM are set to $64$, i.e., $m=d=d_Q=d_K=d_v=64$ while the number of heads of the attention module is set to $2$. The ANP module uses the hidden size $64$ in the encoder and the decoder. The main part of ANP is borrowed from open-source code of DeepMind \emph{https://github.com/deepmind/neural-processes}. The RL module uses the discount factor $\gamma=0.95$ and hidden size $64$ in the policy network. Durign training, we set batch size $128$, use early stopping and weight decay for regularization and apply Adam \citep{kingma-15} to optimize the model. A sample code can be found at \it{https://github.com} \it{/iLevyFan/meta$\_$rl$\_$scaling}.

\subsubsection{Environment} All experiments run on a Linux server (Ubuntu 16.04) with Intel(R) Xeon(R) Silver 4214 2.20GHz CPU, 16GB memory, with a V100 GPU.

\end{document}